\documentclass[11pt,a4paper]{article}
\usepackage[hyperref]{acl2020}
\usepackage{times}
\usepackage{latexsym}

\usepackage{microtype}
\usepackage{graphicx}
\usepackage{subfigure}
\usepackage{booktabs} % for professional tables
\usepackage{multirow}
\usepackage{hhline}
\usepackage{array}
\usepackage{colortbl} 
\usepackage{amsmath, amssymb}
\usepackage{float}

\aclfinalcopy % Uncomment this line for the final submission
\title{Generative Adversarial Networks for Annotated Data Augmentation in Data Sparse NLU}

\author{Olga Golovneva \\
  Amazon / Cambridge, MA \\
  \texttt{olggol@amazon.com} \\\And
  Charith Peris \\
  Amazon / Cambridge, MA \\
  \texttt{perisc@amazon.com} \\}

\date{}

\begin{document}
\maketitle
\begin{abstract}
Data sparsity is one of the key challenges associated with model development in Natural Language Understanding (NLU) for conversational agents. The challenge is made more complex by the demand for high quality annotated utterances commonly required for supervised learning, usually resulting in weeks of manual labor and high cost. In this paper, we present our results on boosting NLU model performance through training data augmentation using a sequential generative adversarial network (GAN). We explore data generation in the context of two tasks, the bootstrapping of a new language and the handling of low resource features. For both tasks we explore three sequential GAN architectures, one with a token-level reward function, another with our own implementation of a token-level Monte Carlo rollout reward, and a third with sentence-level reward. We evaluate the performance of these feedback models across several sampling methodologies and compare our results to upsampling the original data to the same scale. We further improve the GAN model performance through the transfer learning of the pre-trained embeddings. Our experiments reveal synthetic data generated using the sequential generative adversarial network provides significant performance boosts across multiple metrics and can be a major benefit to the NLU tasks.
\end{abstract}

\section{Introduction}
Over recent years, various task-oriented conversational agents, such as Amazon Alexa, Apple’s Siri, Google Assistant, and Microsoft’s Cortana, have become more popular in people’s everyday life and are expected to be highly intelligent. For the NLU component, this means that we expect models to perform recognition of the actions and entities within a user’s request with high accuracy. When first training an NLU model on a new language (a process referred to as bootstrapping a new language), there is a strong requirement for high quality annotated data that would support the most common user requests across a range of domains. As the modeling space expands to support new features and additional languages, NLU models are regularly re-trained on updated data sets to ensure support for these new functions. The major bottleneck in both of these processes is the labor and cost associated with collecting and annotating new training utterances for every new feature or language. 

Recent advances in machine learning methods, including the use of techniques such as transfer learning~\citep{Lu2015TransferLU} and active learning~\cite{Settles2009ActiveLL}, can lead to more efficient data usage by NLU models and therefore decrease the need for annotated training data. Additionally, data augmentation models are being widely explored. The advantage of data augmentation is that once synthetic data is generated, it can be ingested into subsequent models without additional effort, allowing for faster experimentation. 

NLU models in dialog systems can perform a variety of tasks~\cite{Ram2018ConversationalAT, Gao2018NeuralMD}. In this study, we will focus on three of them: \textbf{Domain classification (DC)} -- identify the domain that the user request belongs to (music, reminders, alarm, etc.), \textbf{Intent classification (IC)} -- extract actions requested by users (play music, find a restaurant, set an alarm, etc.), and \textbf{Named Entity Recognition (NER)} -- identify and extract entities (names, values, dates, locations, etc.) from user requests.

For each utterance we expect our NLU model to output a domain, intent, and set of extracted entities with corresponding tags. For example, if a user requests ``\textit{play Bohemian Rhapsody by Queen}'', we expect the NLU model to return \{\textbf{domain}: \textit{music}, \textbf{intent}: \textit{play\_song}, \textbf{named\_entities}: [(bohemian rhapsody, \textit{song\_name}), (queen, \textit{artist\_name})]\}. We call this output \textit{annotation}, and the utterance along with annotation is called an \textit{annotated utterance}. Named entities with corresponding labels are called \textit{slots}.

For our NLU model to perform well on real-time user requests, we need to train it on a large dataset of diverse annotated utterances. However, there could be some areas of functionality where large datasets for training are not available. To boost model performance in situations where training data is limited, we use synthetic data generated from a small set of unique utterances that cover the basic functionality of the user experience, called \textit{Golden utterances}. We leverage a Sequence Generative Adversarial Networks (SeqGAN) introduced by~\citet{Yu2016SeqGANSG} to generate new utterances from this ``seed'' set, and use these generated utterances to augment training data and evaluate the performance of the classification and recognition tasks. We also investigate how the metrics that we use to evaluate the quality of the generated synthetic data links to the performance boost in the underlying tasks.

\section{Related work}
NLU model boosting through training data augmentation has been an active area of research over the last few years, with more sophisticated techniques and models being developed. Some of these techniques include data resampling, the use of Variational Autoencoders (VAEs) and GANs. \citet{Xie2017DataNA} generalize resampling methods by proposing noising schemes that are designed to smooth input data by randomly changing the word tokens in a sentence. First described by~\citet{Kingma2013AutoEncodingVB}, VAEs learn distributed representations of latent variables, and decode random samples to generate data that have similar characteristics to those that the network was trained on. GAN model proposed by~\citet{Goodfellow2014GenerativeAN} includes two competing neural networks: a generator that creates fake data, and a discriminator that is trained to distinguish between fake and real data. The generator is trained on the results of its success in fooling the discriminator and this contest results in synthetic data that is progressively more similar to real data.

Synthetic data have shown to be useful for IC model boosting. For example, \citet{Malandrakis2019ControlledTG} explored a set of encoder-decoder models and proposed the use of conditional VAEs (CVAEs) to generate phrase templates, called carrier phrases. Authors used CVAEs to control the domain, intent, and slot types to generate desirable outputs that resulted in a higher F1 score on the intent classification task.

\citet{Kumar2019ACL} focused on a few-shot IC problem where new categories with limited training data are introduced into an existing system with mature categories. They compared different techniques that were designed to augment training data, including upsampling, random perturbation, extrapolation, CVAEs, and delta-encoders, and combined feature space augmentation with popular BERT pre-training~\cite{Devlin2019BERTPO} to provide better performance.

The use of GANs has been previously explored for text data augmentation in language modeling ~\cite{Kusner2016GANSFS, Yu2016SeqGANSG, Che2017MaximumLikelihoodAD, Guo2018LongTG, Hu2017TowardCG, Li2017AdversarialLF, Lin2017AdversarialRF, Zhang2017AdversarialFM, Fedus2018MaskGANBT} and sentiment classification~\cite{Gupta2019DataAF}. However, discrete text sequence generation brings about several challenges: first, one needs to generate a set of discrete tokens from a random sample of real-valued continuous data, and second, GANs are designed to give feedback on entire sequences, whereas generators need guidance for each subsequent token. The SeqGAN model developed by~\citet{Yu2016SeqGANSG} attempts to resolve these issues by applying reinforcement algorithms for the GAN objective with a policy gradient that evaluates current state-action value using Monte Carlo (MC) search. In this work, we adopt a SeqGAN model to boost DC, IC, and NER tasks in NLU models that suffer from sparse data limitations.

\section{Methods}
\subsection{Data}
For our experiments, we used the English data\footnote{https://fb.me/multilingual\_task\_oriented\_data} collected by \citet{Schuster2019CrosslingualTL}. This data was consisted of three domains: \textit{weather}, \textit{alarm}, and \textit{reminder}, and a total of 43000 utterances. It was collected in a three-step process: step 1 consisted of native English speakers producing utterances for each intent, step 2 consisted of two annotators labeling the intents and slots while any conflicts between these two annotators were resolved in step 3 by a third annotator. The data was processed further to match the format that was suitable for our models. 

\subsection{Models}
Text data boosting in NLU was extensively used for classification tasks, so most previous research focused on generating sentences~\cite{Kumar2019ACL}, carrier phrases~\cite{Malandrakis2019ControlledTG}, or embeddings~\cite{Guo2018LongTG}. In our work we consider DC, IC, and NER problems, where both sequence and word tags are needed for model training. We leverage a GAN to synthesize training data as a sequence of intents and slots: $X= \{x_0,x_1,…,x_n\}$, where $n$ varies between a length of $1$ and the maximum allowed utterance length. Each slot $x_i  (i>0)$ denotes the combination of the $i^{th}$ word and its corresponding tag, and $x_0$ is a concatenation of the utterance domain and intent. For example, for the utterance ``\textit{play Bohemian Rhapsody by Queen}'', the training sequence for text generation would be as follows: ``\textit{music/play\_song play:none bohemian:song\_name rhapsody:song\_name by:none queen:artist\_name}''.

\subsubsection{SeqGAN model}
When applied to text data, traditional GANs have difficulty performing back-propagation due to the non-differentiable output of the generator model. The SeqGAN model addresses this issue by treating the generator as a reinforcement learning agent that optimizes the GAN objective. The discriminator itself is used within the reward function to evaluate output sequences and return feedback to guide the learning of the generative model. Traditional reward functions from classification models are also limited by the ability to only provide score/loss-based reward values for a complete sequence. The SeqGAN model enables evaluation of the action-value for an intermediate state of unfinished sequence for each initial state, $s_0$, by applying an MC search with a rollout policy to sample the unknown last tokens~\cite{Yu2016SeqGANSG}. MC search is a tree-search algorithm with a root node $s_0$. Each child node of the tree is drawn from a distribution parametrized by a stochastic parametrized policy. This policy can be any, for example current state of the generator $G_\theta$.

\begin{figure*}[ht]
\begin{center}
\centerline{\includegraphics[width=0.7\textwidth]{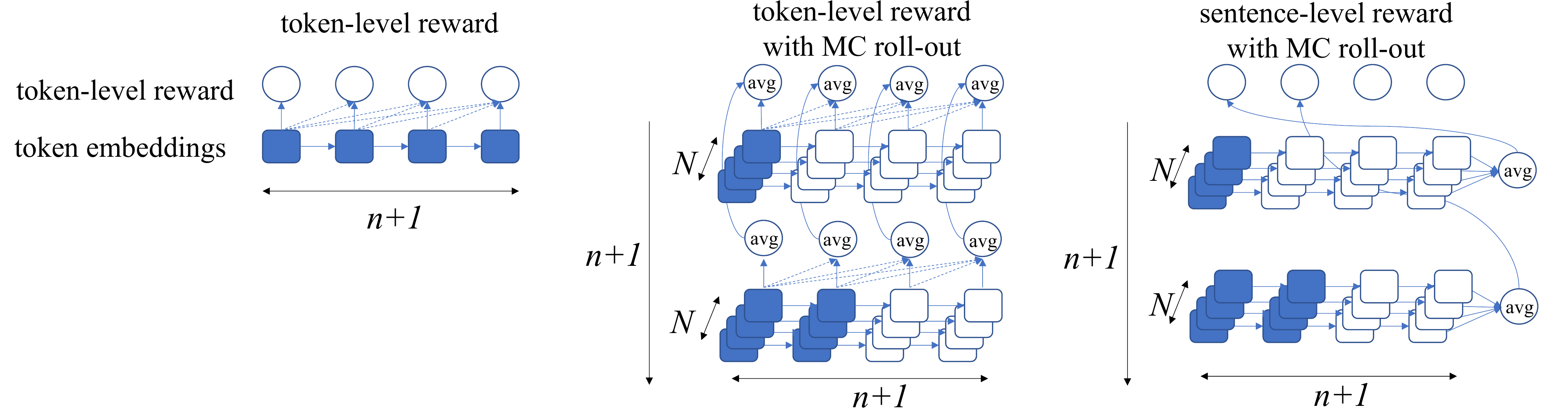}}
\caption{Schematic representation of the reward function produced by discriminator model for each of the $n+1$ input tokens and N Monte-Carlo rollouts {\it (avg = average)}. Tokens fixed at each iteration are represented by solid squares and tokens produced by the rollout policy are represented by blank squares. Sentence-level reward produces feedback on a complete sentence only, whereas token-level reward gives feedback on each token conditioned on previous set of tokens.}
\label{reward}
\end{center}
\end{figure*}

In our experiments, we compare different ways to compute the reward function (Figure~\ref{reward}). First, we use an implementation where the MC tree search strategy is replaced with token-level reward produced by the discriminator~\cite{Xu2018DiversityPromotingGA, Hu2018TexarAM}. We use the output of the Long Short-Term Memory (LSTM)-based discriminator model $D_\phi^L$, cross-entropy, as the reward. For a synthetic sentence $Y=\{y_0,y_1, ...,y_n\}, y_i\in \mathcal{Y}$, where $\mathcal{Y}$ is the vocabulary of candidate domain-intent and token-label pairs, the cross-entropy based reward for the $i^{th}$ word is calculated as:
\begin{equation}
\label{token-lvl}
R(y_i) = -\log D_\phi^L(y_i|y_{<i})
\end{equation}

Next, we devise our own MC search-based method to produce a set of possible sequences to approximate the expectation of the token-level reward, rather than using a single evaluation of only one of the possible sequences. Using the current state of the generator model $G_\theta$ as a stochastic rollout policy, for each incomplete sequence $\{y_0,y_1,...,y_k\}$ we use Monte-Carlo search to produce $N$ complete sequences. We evaluate the token-level reward for each of these sequences:
\begin{equation}
\label{MC}
\{y_{k+1}, ..., y_n\} \in MC^{G_\theta}(\{y_0, ..., y_k\}; N),
\end{equation}
for ${k<n}$. To approximate the expected value of the reward function we average the token-level reward (\ref{token-lvl}) calculated for the $N$ MC rollouts:
\begin{align}
\label{ours}
&R(y_i) = \nonumber\\
&\;\;\;\;\frac{1}{(n+1)N}\sum_{k=0}^{n}\sum_{j=1}^{N}(-\log D_\phi^L(y_i|y_{<i}))_{j,k},
\end{align}
where $(...)_{j,k}$ is the $j_{th}$ MC rollout of the $k_{th}$ sequence. Finally, we compare results with an expectation of a sentence-level reward calculated using MC tree rollout strategy for the convolutional neural network (CNN)-based discriminator feedback $D_\phi^C$, that provides feedback on the full sentence~\cite{Yu2016SeqGANSG}. Intermediate token-level reward for current token $y_i$  is approximated using MC rollouts as follows:  
\begin{align}
\label{MCR}
R_{MC}(y_i) = \frac{1}{N}\sum_{j=1}^{N}(D_\phi^C(Y))_j, i<n, 
\end{align}
where the sampling of the missing $(n-i)$ tokens is governed by the equation~(\ref{MC}). At each step $i$, generator reward is governed by the Monte-Carlo approximation~(\ref{MCR}), or discriminator feedback on the full sentence:
\begin{align}
\label{seqgan}
R(y_i) = \begin{cases}
	      R_{MC}(y_i), i<n, \\
	      D_\phi^C(Y), i = n.
	\end{cases} 
\end{align}
The generator is trained to maximize the reward from the discriminator model for the full sequence~\cite{Sutton1999PolicyGM}:
\begin{align}
&J(\theta)=\mathbb{E}[R(Y|s_0, \theta)]=\nonumber\\
&\;\;\;\;\;\;\;\;\;\;\;\;\;\;\sum_{y_0\in \mathcal{Y}}G_\theta(y_0|s_0)Q_{D_\phi}^{G_\theta}(s_0, y_0)
\end{align}
where $Q_{D_\phi}^{G_\theta}(s_0, y_0)$ can be estimated by $R(y_0)$ using equations~(\ref{token-lvl}),~(\ref{ours}) or~(\ref{seqgan}). 

The pre-training procedure for the generative decoder uses the maximum likelihood estimation metric, followed by the pre-training of the discriminative classifier on positive samples from the training data and negative samples produced by the generator. After pre-training, the generative and discriminative models are trained one at a time in the following loop: the last discriminator state is used to provide feedback for the generative model training, which then provides a new set of negative examples for the discriminator updates.

\subsubsection{DC and IC/NER models}
\begin{figure}[ht]
\begin{center}
\centerline{\includegraphics[width=0.8\columnwidth]{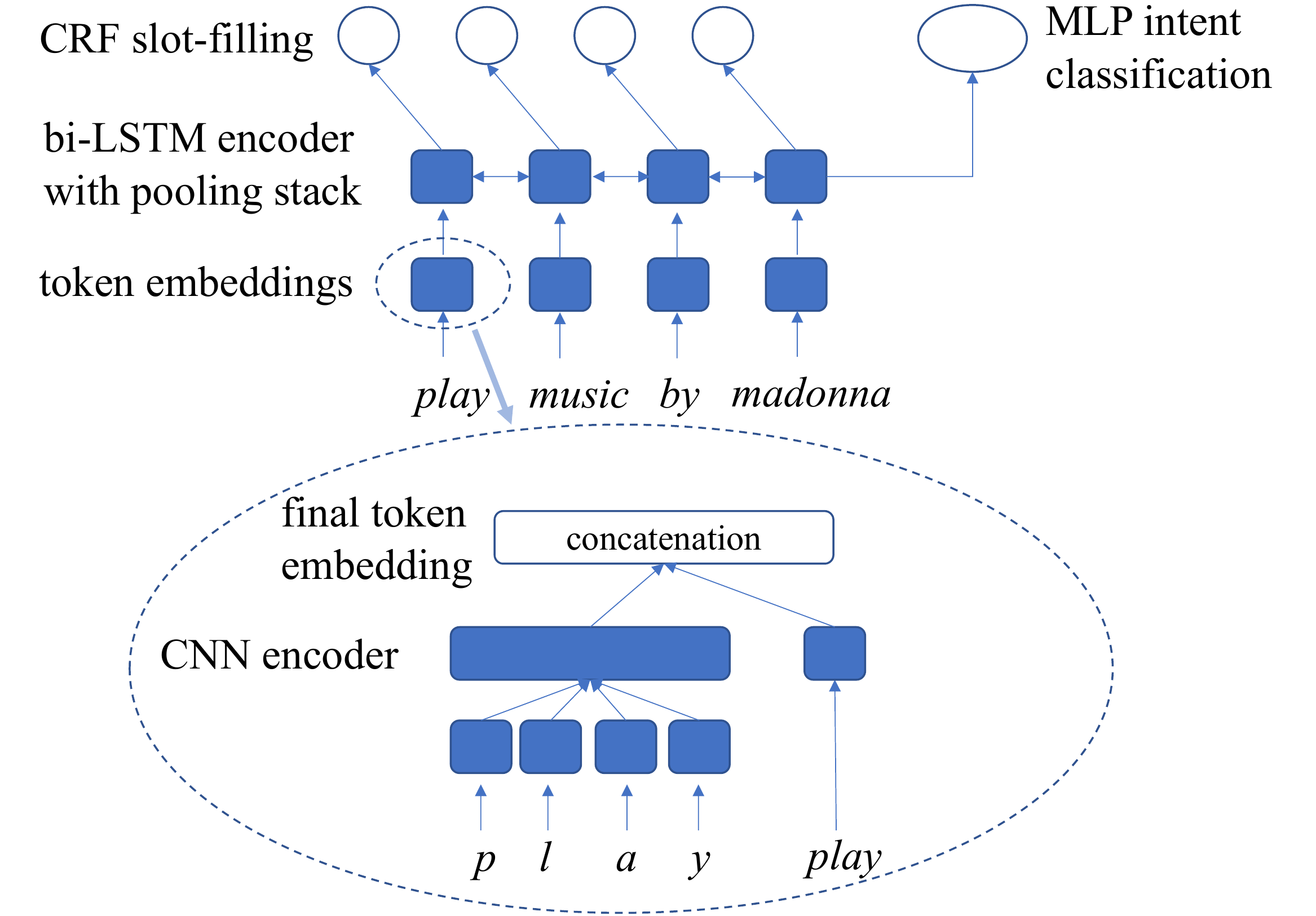}}
\caption{Multi-task model for IC and NER prediction.}
\label{multi-task}
\end{center}
\end{figure}

Our model architecture consists of two parts: a sentence classification model for the DC task, and a domain-specific joint model for the IC and NER tasks. The DC model is the same as the IC part of the joint IC-NER model, so we will describe only the latter in detail.
The IC-NER model components are schematically shown in Figure~\ref{multi-task}, and are composed of the following:
\begin{itemize}
\item \textbf{Embedding}: concatenation of word embedding with 256 dimensions and character embedding with 16 dimensions trained on a 1-filter CNN with a \textit{tanh} activation function and dropout. 
\item \textbf{Encoder}: 2-layer bidirectional LSTM with 384 dimensions in each hidden layer with dropout and layer normalization for token encoder, and a pooling stack for sequence encoder.
\item \textbf{Decoders}: MLP classifier with 256 dimensions for IC task and CRF sequence labeler with 192 dimensions. For each block we also apply ELU activation function and drop-out. All IC-NER models were trained for 500 epochs, and DC models for 100 epochs.  
\end{itemize}

\subsection{Experimental setup}
\begin{figure}[ht]
\begin{center}
\centerline{\includegraphics[width=0.95\columnwidth]{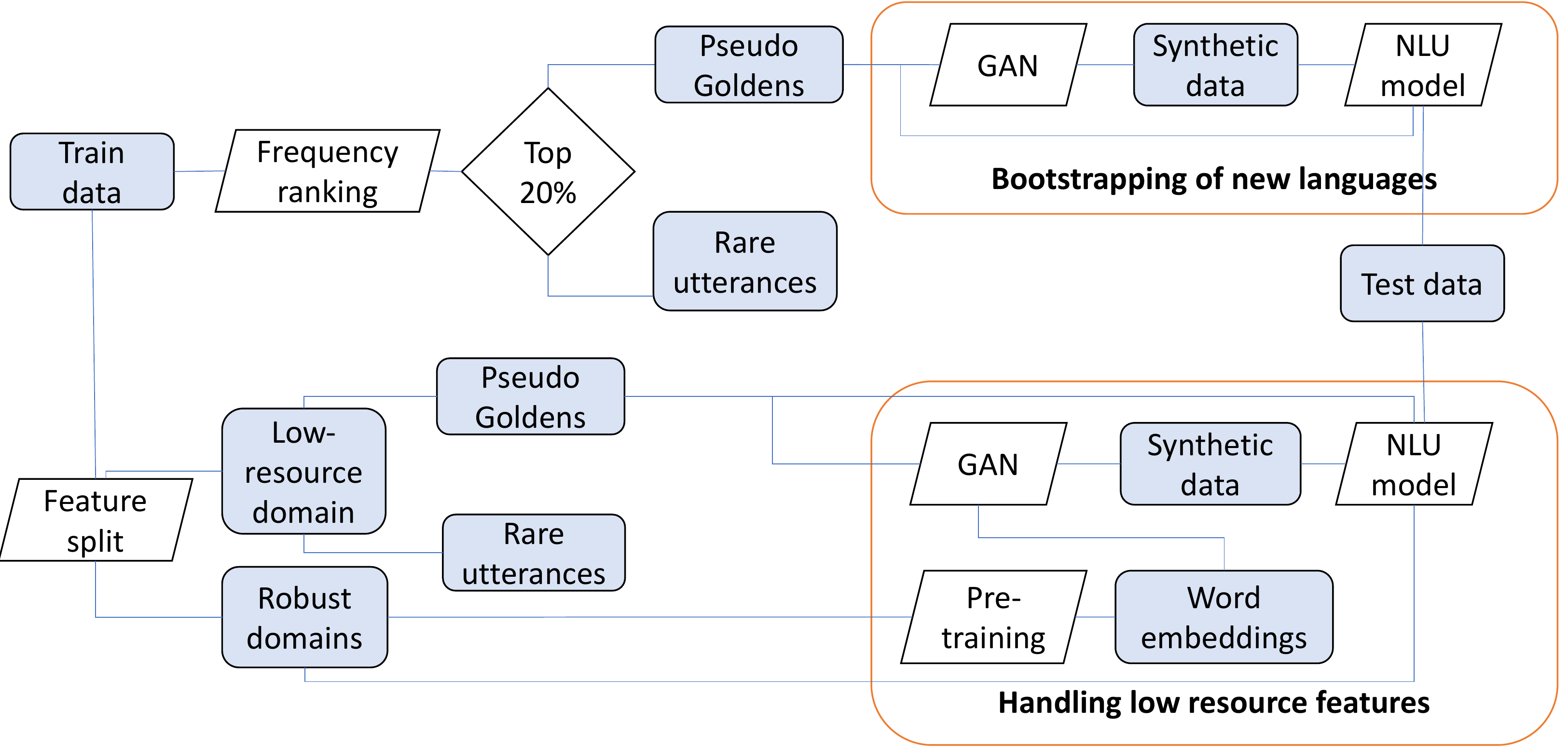}}
\caption{This schematic represents our data and experimental process. For our experiments we use annotated data collected by \citet{Schuster2019CrosslingualTL}. Train and validation sets were combined together in a single train data set to extract pseudo Golden utterances. Pseudo Goldens were further used to train and evaluate the GAN model. The test set was untouched and was used to evaluate model performance in full scale. To simulate the low resource feature task, we used the full train data set for the two robust domains, and the pseudo Goldens to represent the third low-resource domain.}
\label{exper}
\end{center}
\end{figure}

We build our experiments to mimic two major tasks that are of interest to us (Figure~\ref{exper}):
\begin{enumerate}
\item Bootstrapping of a new language 
\item Handling low resource features 
\end{enumerate}

We manipulate our English data set to simulate modeling conditions in situations where training data is limited. For our tasks, we need to identify a set of utterances from within our data set that can be classified as Golden utterances, i.e., utterances that are found to be usually common across languages and cover the basic functionality of the user experience. To select Goldens from our data set we use the following process. We sort the utterances in our training data by the utterance frequency within each domain-intent combination. We then select the top 20\% of utterances within each domain-intent combination, unique them, and call them \textit{pseudo Goldens}. The unique utterances within the remaining set are called \textit{Rare} utterances. Utterance counts are provided in Appendix~\ref{tab:appa}.

\subsubsection{Bootstrapping NLU models for new languages}\label{bootstr}
When bootstrapping an NLU model to support a new language for a conversational artificial intelligence (AI), there is usually a very limited data available. One major data source is Goldens translated from another existing robust language model. In mimicking this task, we use our determined pseudo Goldens as input for both data generation and NLU models. 
Synthetic data generated with the GAN model is added to the set of pseudo Goldens and fed into our NLU model. The NLU models are tested using our fixed original test data set (Figure~\ref{exper}). Our baseline for comparison consists of an NLU model run on only the pseudo Goldens.

\subsubsection{Handling low resource features}\label{lowres}
Another common challenge faced by conversational AI NLU models is sparse data. When a feature is new, or not common, we do not have enough data for the NLU model to generalize well on possible request variations. But, unlike the previous task, we typically do have a large amount of data collected in the same language, but within different intents and domains. Generative adversarial networks, paired with a transfer learning approach, can give us the opportunity to use other domains and intents which have robust data, to strengthen the performance of these low resource features. 
In this case we limit our experiments to exploring a case where we have a low resource domain. In mimicking this task, we pick one of the three domains we have as the low resource domain. We consider only the pseudo Goldens as available data for this domain. The remaining two domains are considered as robust domains and we consider all data (i.e., both pseudo Goldens and Rare utterances) as available data for the NLU model. We also use all training data from the two robust domains to pre-train word embedding using the fastText algorithm~\cite{Lample2017UnsupervisedMT}. We then run data generation for the low resource domain using our SeqGAN implementation while using the pre-trained word embeddings from the two robust domains to initialize word embeddings both in generator and discriminator. For the NLU model, we feed the pseudo Goldens together with the synthesized data for the low resource domain and feed all available training data for the two robust domains. We measure the performance of the low resource domain against a baseline which is run using only pseudo Goldens from the low resource domain together with all available data from the robust domains. As with task 1, performance is tested on the separate fixed test data set.

\section{Data Generation}
We explore the following different GAN frameworks and expansions to generate synthetic annotated utterances:
\begin{itemize}
\item \textbf{Original implementation with Monte Carlo rollout}: a selected set of experiments to benchmark the original implementation by~\citet{Yu2016SeqGANSG} on our open source data set.  
\item \textbf{Original implementation without rollouts}: a selected set of experiments to benchmark the original implementation by~\citet{Yu2016SeqGANSG}, where the reward function is evaluated on a single MC rollout. 
\item \textbf{Generator with token-level reward}: training the generator using token-level reward as suggested by~\citet{Hu2018TexarAM}.  
\item \textbf{Generator with token-level Monte Carlo rollout}: we expand the above implementation to include a token-level Monte Carlo rollout and test our task of bootstrapping new languages.  
\item \textbf{Generator with pre-trained embeddings}: we add fastText pre-trained embeddings to the generator and discriminator to selectively test our task of handling low-resource features).  
\end{itemize}
The model architecture consists of three parts: generator pre-training, discriminator pre-training, and adversarial training. In the pre-training parts, we first train the generator, followed by the discriminator, for 80 epochs each. For each adversarial epoch, we update the generator once and then for 35 steps we generate negative examples using the current state of the generator, combined with the same number of positive examples from the training data, and re-train the discriminator. The total number of adversarial training epochs is set to 600. All hyper-parameters were chosen based on observations of when the loss functions and synthetic data quality evaluations either stabilize or clearly degrade.

\section{Results and Discussion}
\subsection{Evaluation}
We evaluate our models using domain accuracy, intent accuracy, slot F1, and frame accuracy. Domain and intent accuracy measure the accuracy of the domain and intent classification tasks, respectively. We use micro-averaging to calculate slot F1 to measure the performance of the NER task. Frame accuracy indicates the relative number of utterances for which the domain, intent, and all slots were correctly identified. In our case, we pay attention to individual metrics to understand which tasks are most affected by the synthesized data.
\subsection{Language bootstrapping task}
In this section we present the results of our experiment mimicking the task of bootstrapping a new language. See Section \ref{bootstr} for experimental setup. We compare three SeqGAN implementations each with a different reward policy. 
An open question when using generated data for NLU model training is whether the improvements observed are due to the data enrichment gained by the new variations introduced in synthetic data or due to upsampling. To test this, we repeat each experiment on three subsamples of the generated synthetic data. First, we add a synthetic data set that is equal in size to the pseudo Golden data set. We call this \textit{TopX} sampling. This enables us to explore the changes in performance obtained by adding a synthetic data set that reflects the original distribution produced by the trained GAN framework, but with limited effects of upsampling. In the second case, we produce a synthetic data set that is significantly larger than the pseudo Golden data set (9600 utterances per domain) and we take only the unique utterances within that set. We call this \textit{Uniques} sampling. This enables us to explore the changes in performance obtained by adding a synthetic dataset that contains an exhaustive set of the different combinations of utterances that the GAN can create, given the input data. Finally, we add the full set of 9600 generated utterances per domain, that should be similar in distribution to the pseudo Goldens set (\textit{All} sampling). The results obtained for each SeqGAN implementation with TopX, Uniques, and All sampling strategies are summarized in Appendix~\ref{tab:appb1} and Appendix~\ref{tab:appb2}. 

\begin{table*}[h]
\caption{Performance relative to the baseline for models with token-level and sentence-level reward using All sampling strategy. Baseline is a model trained on Goldens only.}
\label{no-mc}
\centering
\begin{small}
\setlength\extrarowheight{1.5pt}
\begin{tabular}{p{1.5cm}p{1cm}p{0.98cm}p{1cm}p{1cm}p{1.1cm}p{0.98cm}p{1cm}p{1cm}p{1cm}p{1cm}}
\hline
{\textbf{Reward}} & {\textbf{Domain}} & \multicolumn{3}{c}{\textbf{Intent accuracy}} & {\textbf{Overall}} & \multicolumn{3}{c}{\textbf{Slot F1}} & {\textbf{Overall}} & {\textbf{Frame}}\\
\cline{3-5}\cline{7-9}
 {}& {\textbf{accuracy}}& {\textbf{alarm}} & \textbf{reminder} & \textbf{weather} & {\textbf{int. acc.}} & \textbf{alarm} & \textbf{reminder} & \textbf{weather} & {\textbf{slot F1}}& {\textbf{accuracy}}\\ \hline
Token-level & 0.29 & 11.84 & 14.62 & -3.89 & 4.41 & -1.04 & 0.8 & 1.25 & 0.39 & 3.26 \\ \hline
Sent.-level & -0.26 & 14.45 & 17.4 & 0.27 & 7.81 & -1.19 & 1.21 & -2.77 & -1.23 & -4.43 \\ \hline
\end{tabular}
\end{small}
\end{table*}

\begin{table*}[h]
\caption{Performance relative to the baseline for models with Monte Carlo rollouts on token-level and sentence-level reward using Uniques sampling strategy. Baseline is a model trained on Goldens only.}
\label{mc}
\centering
\begin{small}
\setlength\extrarowheight{1.5pt}
\begin{tabular}{p{1.5cm}p{1cm}p{0.98cm}p{1cm}p{1cm}p{1.1cm}p{0.98cm}p{1cm}p{1cm}p{1cm}p{1cm}}
\hline
{\textbf{Reward}} & {\textbf{Domain}} & \multicolumn{3}{c}{\textbf{Intent accuracy}} & {\textbf{Overall}} & \multicolumn{3}{c}{\textbf{Slot F1}} & {\textbf{Overall}} & {\textbf{Frame}}\\
\cline{3-5}\cline{7-9}
 {}& {\textbf{accuracy}}& {\textbf{alarm}} & \textbf{reminder} & \textbf{weather} & {\textbf{int. acc.}} & \textbf{alarm} & \textbf{reminder} & \textbf{weather} & {\textbf{slot F1}}& {\textbf{accuracy}}\\ \hline
Token-level & 0.41 & 14.83 & 14.43 & -0.12 & 7.21 & -0.43 & 2.82 & 0.25 & 0.67 & 4.3 \\ \hline
Sent.-level & -1.37 & 9.07 & 9.19 & 0.23 & 4.71 & -0.09 & 1.47 & -0.52 & -0.12 & 0.19 \\ \hline
\end{tabular}
\end{small}
\end{table*}

\subsubsection{Generator with token-level reward}
When using the Generator with token-level reward, we observe that out of three synthetic data sampling strategies (i.e., TopX, Uniques, and All), the biggest overall gain is shown by the All sampling strategy with an overall intent accuracy improvement of ~4\%, and an overall frame accuracy improvement of ~3\% (Table~\ref{no-mc}). Domain accuracy and overall slot F1 do not show much change. In this setup, alarm and reminder domains’ intent accuracies show increases of 12-14\% with overall intent accuracy increasing by ~4\%. However, intent accuracy in the weather domain degrades by ~4\%. The generator with token-level reward outperforms the corresponding generator with sentence-level reward in domain accuracy, slot F1 and frame accuracy, but trails in intent metrics.

When the models are run using Goldens upsampled to the same counts, we observe that overall domain, intent, and frame accuracies are slightly better. However, individual intent accuracy of alarm and reminder domains do not perform as well as with synthetic data. No degradation of the weather domain is observed with upsampled data.

\subsubsection{Generator with token-level Monte Carlo rollout}
Using a Generator with token-level Monte Carlo rollout brings significant improvement, especially outperforming other models in the Uniques sampling strategy. Overall intent accuracy shows an improvement of ~7\%, and overall frame accuracy improves by ~4\% (Table~\ref{mc}).  Alarm and reminder domains' intent accuracy show improvements of ~14-15\%. The reminder domain’s slot F1 improves by ~3\%. This setup performs significantly better than the generator with sentence-level reward in all overall metrics and also in alarm and reminder domain intent accuracies.

When compared to the models run using Goldens upsampled to the same counts, we observe that the Generator with token-level Monte Carlo rollout policy performs better on domain accuracy, overall intent accuracy, and overall slot F1 while being slightly under on frame accuracy. Specifically, it performs much better on both the alarm and reminder domains which show approximately 50\% the performance boost.

\subsection{Handling low resource features}
In this section we present the results of our experiment mimicking the task of handling low resource features. See Section~\ref{lowres} for the experimental setup. We conduct each experiment three times and present the mean results for Uniques sampling strategy in Appendix~\ref{tab:appc}. 

\subsubsection{Generator with token-level reward and embeddings pre-trained on robust domains}
Appendix~\ref{tab:appc} summarizes the results obtained using SeqGAN with a generator with token-level reward and embeddings pre-trained on robust domains to synthesize data for the low-resource domain. We observe a 12\% increase in intent accuracy in the alarm domain when compared to the baseline. We also observe a 13\% increase in intent accuracy in the reminder domain when synthetic data is added. For the weather domain, we do not observe a significant change in intent accuracy.
For alarm and reminder domains we see an increase in overall intent accuracy when synthetic data is added. For these same domains, overall frame accuracy shows small improvements while domain accuracy and overall slot F1 does not show any significant changes.

\subsubsection{Generator with token-level Monte Carlo rollout and embeddings pre-trained on robust domains}
Appendix~\ref{tab:appc} shows the results obtained when using MC rollout policy in addition to using embeddings pre-trained on robust domains to synthesize data for the low-resource domain. For the reminder and alarm domains, we note that the performance boost in the IC task for the low-resource domain is larger by 2-3\% than without the MC rollout. The weather domain shows a small but statistically significant improvement of 0.5\% when compared to the baseline. These results suggest that the MC rollout policy provides additional guidance to the generator in all cases.

\section{Synthetic data deep dive}
\subsection{Evaluating the quality of the data generated}
To the best of our knowledge, there is no comprehensive metric that is commonly used for measuring the performance of a text generation model. To measure the performance of the SeqGAN, we use the n-gram Bilingual Evaluation Understudy (BLEU) score~\cite{Papineni2001BleuAM}, calculated against a test set of Golden utterances. This metric measures the degree of similarity between the generated data and test set. Additionally, we calculate the diversity of generated data through the number of unique phrases we generated and number of unique words used from vocabulary. We also keep track on the mean utterance length to detect possible utterance collapse for token-level feedback, when the generator learns to generate shorter utterances. Detailed evaluations on bootstrapping experiments are provided in the Appendix~\ref{tab:appd}, and summarized in Figure~\ref{bleuscores}. We see that compared to other domains, weather’s BLEU scores are higher, and the diversity of the generated data is lower, suggesting that the model potentially reproduces almost the same data as the input, and does not bring in much novelty for NLU model training.

\begin{figure}%*}[h]
%\vskip 0.2in
\begin{center}
\centerline{\includegraphics[width=0.5\textwidth]{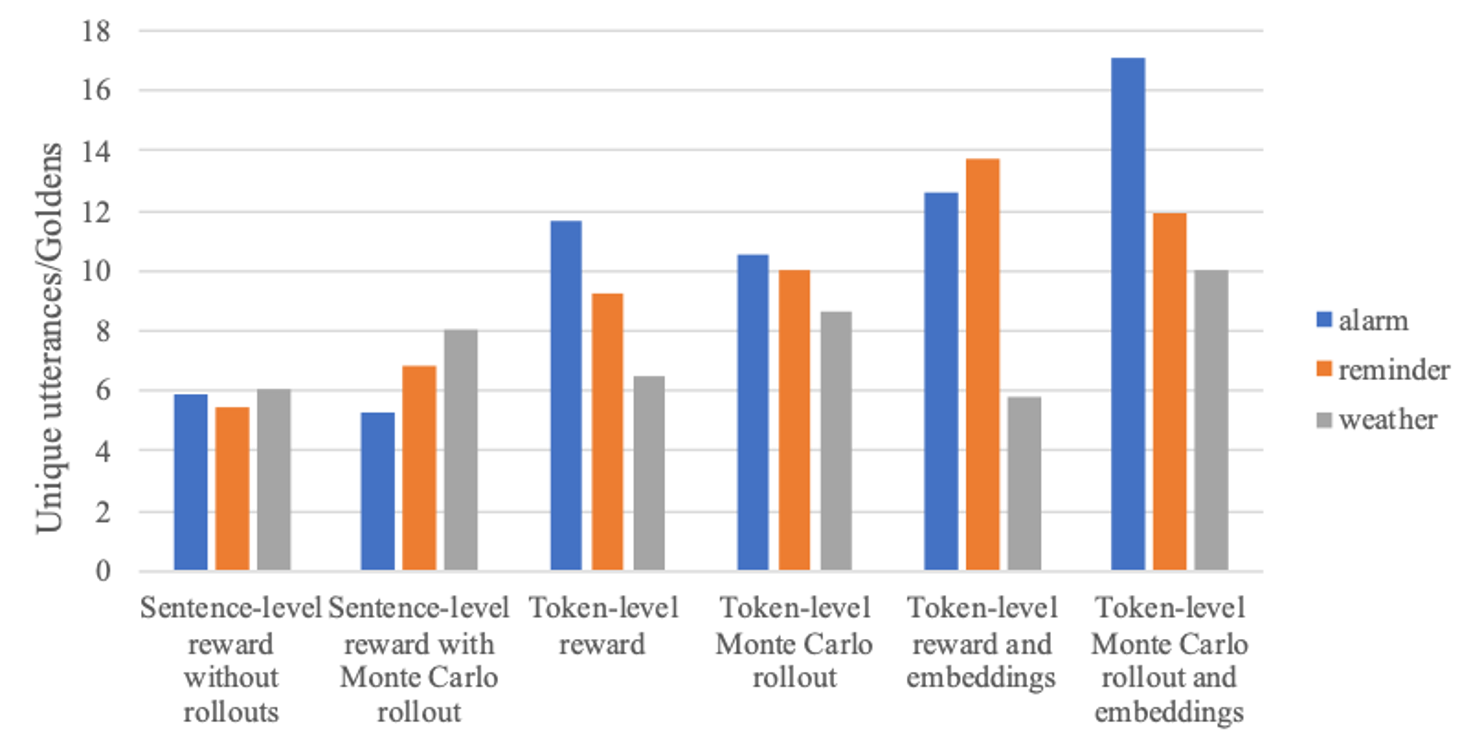}}
\centerline{\includegraphics[width=0.5\textwidth]{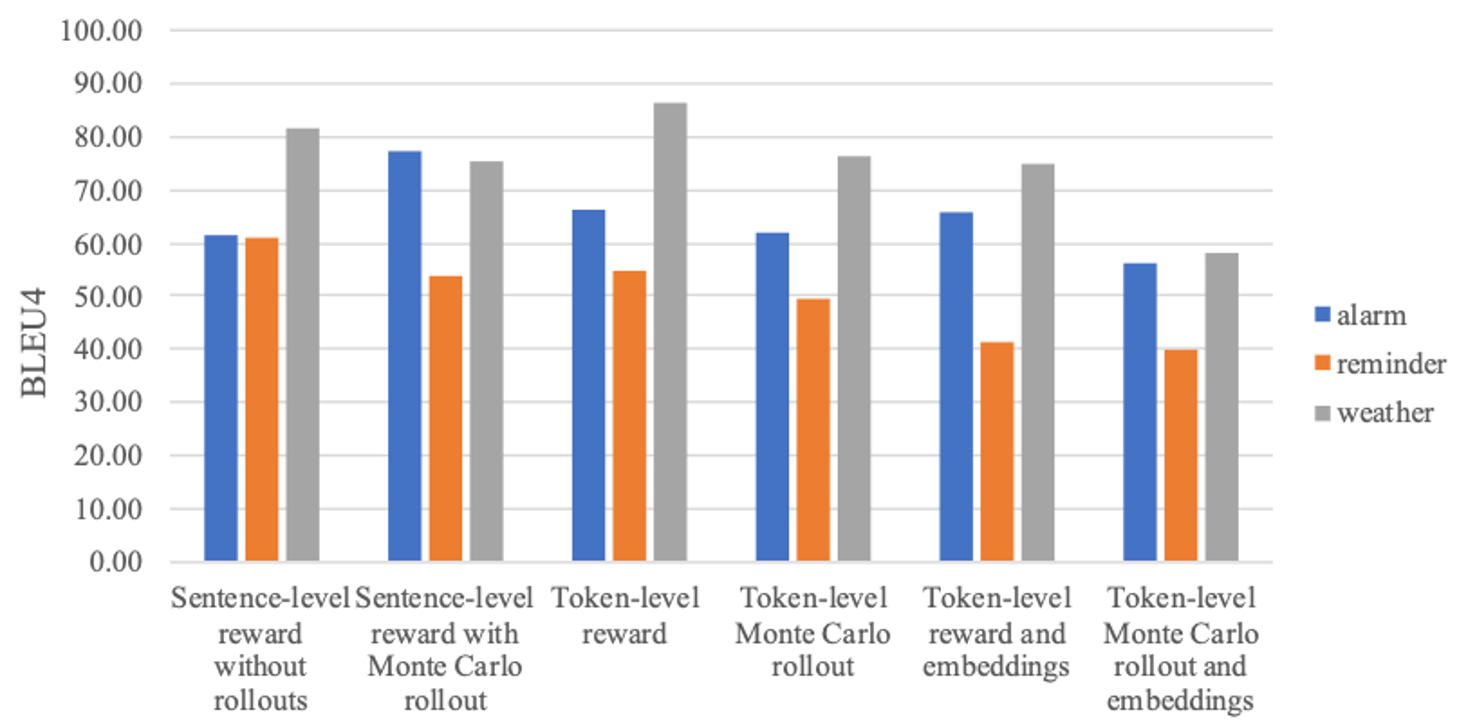}}
\caption{Data quality evaluation for synthetic data generated in different GAN models: number of unique utterances generated normalized by the number of Goldens used as an input for GAN model (left), and 4-gram BLEU score (right).}
\label{bleuscores}
\end{center}
%\vskip -0.2in
\end{figure}%*}

\subsection{Annotation review}
Although, in general, we observed that the NLU model benefits from synthetic data, we have noted some degradation in NER models, especially in the weather domain. Deep diving into NER errors, we found the following major sources of errors: annotation errors and context-dependent annotations. First, in seeding pseudo Goldens, we selected utterances based on their frequency, and then took all unique annotations to be a pseudo Goldens set. That process artificially increased weight of annotation errors in cases were for frequent phrases there were a few misannotated utterances. For example, for utterances ``\textit{remind me to ... tomorrow}'' with and without label ``\textit{datetime}'' frequently appeared in our goldens, and the NLU model fails to recognize ``\textit{tomorrow}'' as an entity. In contrast, for the utterances ``\textit{remind me tomorrow to ...}'' the model produces the correct ``\textit{datetime}'' label. Additionally, small context-dependent words that have different annotation in the same domain, but appear to have a dominant annotation in pseudo Goldens and further in synthetic data, happen to be another cause of failures. One example of such a word is ``\textit{for}'' in the weather domain, where in ``\textit{weather for London next week}'' it has no label, while in ``\textit{forecast for next week please}'' it is labeled as datetime. 

\section{Conclusions}
In this paper, we evaluate the use of the SeqGAN model for synthetic annotated data generation to boost NLU model performance. We have shown that adding synthetic data to bolster our Goldens can significantly improve DNN model performance in intent classification and named entity recognition tasks. We propose a token-level reward with Monte Carlo search rollout to guide the generator model, that showed better performance when compared with a regular token-level reward implementation, sentence-level reward implementations both with and without Monte Carlo tree search, and with a pure upsampling strategy.
We also show that using SeqGAN together with embeddings pre-trained on high-resource domains to generate synthetic data can significantly improve the performance of low-resource domains. Embeddings pre-trained on different tasks can carry over the information they have learned and that can be especially useful in low-resource model building scenarios.

\bibliography{emnlp2020}
\bibliographystyle{acl_natbib}

\newpage
\onecolumn
\appendix

\section{Unique utterance counts, intent counts and labels counts for pseudo Golden and Rare data. }\label{tab:appa}
\begin{table}[ht]
\centering
%\caption{Unique utterance counts, intent counts and labels counts for pseudo Golden and Rare data. }
\resizebox{13cm}{!}{%
\begin{tabular}{llccc}
\hline
\textbf{Domain}   & \textbf{Utterance Group} & \textbf{Unique Utterance Count} & \textbf{Unique Intent Count} & \textbf{Unique Label Count} \\ \hline
\textbf{alarm}    & \textbf{pseudo Goldens}  & 202                             & 6                            & 3                           \\ \hline
\textbf{alarm}    & \textbf{Rare}            & 7006                            & 6                            & 7                           \\ \hline
\textbf{reminder} & \textbf{pseudo Goldens}  & 513                             & 3                            & 7                           \\ \hline
\textbf{reminder} & \textbf{Rare}            & 5580                            & 3                            & 7                           \\ \hline
\textbf{weather}  & \textbf{pseudo Goldens}  & 228                             & 3                            & 5                           \\ \hline
\textbf{weather}  & \textbf{Rare}            & 10929                           & 3                            & 8                           \\ \hline
\end{tabular}
}
\end{table}

\section{Summary of the results on Language Bootstrapping Experiments averaged over three runs for Generator with sentence-level reward.} \label{tab:appb1}
\begin{table}[ht]
\centering
%\caption{Summary of the results on Language Bootstrapping Experiments averaged over three runs forGenerator with sentence-level reward.}
\resizebox{14cm}{!}{%
\begin{tabular}{llccccccc}
\hline
\multirow{3}{*}{\textbf{Metric}}          & \multirow{3}{*}{\textbf{Domain}} & \multicolumn{1}{c}{\textbf{Baseline}}    & \multicolumn{6}{c}{\textbf{Generator with sentence-level reward with Monte Carlo rollout}}                                                                                                                                                                     \\ \cline{3-9} 
                                          &                                  & \multicolumn{1}{c}{\textbf{}}            & \multicolumn{2}{c}{\textbf{TopX}}                                                  & \multicolumn{2}{c}{\textbf{Unique}}                                                & \multicolumn{2}{c}{\textbf{All}}                                                   \\ \cline{3-9} 
                                          &                                  & \multicolumn{1}{c}{\textbf{Performance}} & \multicolumn{1}{c}{\textbf{Performance}} & \multicolumn{1}{c}{\textbf{\% change}} & \multicolumn{1}{c}{\textbf{Performance}} & \multicolumn{1}{c}{\textbf{\% change}} & \multicolumn{1}{c}{\textbf{Performance}} & \multicolumn{1}{c}{\textbf{\% change}} \\ \hline
\multicolumn{2}{l}{\textbf{Domain accuracy}}                               & 96.83                                     & 97.71                                     & +0.91                                   & 95.5                                      & -1.37                                   & 95.17                                     & -1.72                                   \\ \hline
\multirow{3}{*}{\textbf{Intent accuracy}} & \textbf{alarm}                   & 73.33                                     & 68.53                                     & -6.54                                   & 79.98                                     & +9.07                                   & 80.46                                     & +9.73                                   \\ \cline{2-9} 
                                          & \textbf{reminder}                & 72.93                                     & 81.72                                     & +12.06                                  & 79.63                                     & +9.19                                   & 79.39                                     & +8.85                                   \\ \cline{2-9} 
                                          & \textbf{weather}                 & 97.5                                      & 97.58                                     & +0.08                                   & 97.72                                     & +0.23                                   & 98.65                                     & +1.18                                   \\ \hline
\multicolumn{2}{l}{\textbf{Overall intent accuracy}}                       & 84                                        & 84.89                                     & +1.05                                   & 87.96                                     & +4.71                                   & 88.4                                      & +5.24                                   \\ \hline
\multirow{3}{*}{\textbf{Slot F1}}         & \textbf{alarm}                   & 92.63                                     & 91.8                                      & -0.89                                   & 92.54                                     & -0.09                                   & 93.11                                     & +0.52                                   \\ \cline{2-9} 
                                          & \textbf{reminder}                & 83.49                                     & 84.79                                     & +1.55                                   & 84.72                                     & +1.47                                   & 83.61                                     & +0.14                                   \\ \cline{2-9} 
                                          & \textbf{weather}                 & 83.63                                     & 82.55                                     & -1.29                                   & 83.2                                      & -0.52                                   & 83.63                                     & +0.00                                   \\ \hline
\multicolumn{2}{l}{\textbf{Overall slot F1}}                               & 86.19                                     & 85.88                                     & -0.36                                   & 86.08                                     & -0.12                                   & 86.07                                     & -0.14                                   \\ \hline
\multicolumn{2}{l}{\textbf{Overall frame accuracy}}                        & 46.44                                     & 45.64                                     & -1.73                                   & 46.53                                     & +0.19                                   & 46.51                                     & +0.16                                   \\ \hline
\multirow{3}{*}{\textbf{Metric}}          & \multirow{3}{*}{\textbf{Domain}} & \multicolumn{1}{c}{\textbf{Baseline}}    & \multicolumn{6}{c}{\textbf{Generator with sentence-level reward without rollouts}}                                                                                                                                                                             \\ \cline{3-9} 
                                          &                                  & \multicolumn{1}{c}{\textbf{}}            & \multicolumn{2}{c}{\textbf{TopX}}                                                  & \multicolumn{2}{c}{\textbf{Unique}}                                                & \multicolumn{2}{c}{\textbf{All}}                                                   \\ \cline{3-9} 
                                          &                                  & \multicolumn{1}{c}{\textbf{Performance}} & \multicolumn{1}{c}{\textbf{Performance}} & \multicolumn{1}{c}{\textbf{\% change}} & \multicolumn{1}{c}{\textbf{Performance}} & \multicolumn{1}{c}{\textbf{\% change}} & \multicolumn{1}{c}{\textbf{Performance}} & \multicolumn{1}{c}{\textbf{\% change}} \\ \hline
\multicolumn{2}{l}{\textbf{Domain accuracy}}                               & 96.83                                     & 94.62                                     & -2.28                                   & 97.6                                      & +0.80                                   & 96.58                                     & -0.26                                   \\ \hline
\multirow{3}{*}{\textbf{Intent accuracy}} & \textbf{alarm}                   & 73.33                                     & 73.89                                     & +0.77                                   & 79.77                                     & +8.78                                   & 83.93                                     & +14.45                                  \\ \cline{2-9} 
                                          & \textbf{reminder}                & 72.93                                     & 74.71                                     & +2.44                                   & 87.53                                     & +20.02                                  & 85.62                                     & +17.40                                  \\ \cline{2-9} 
                                          & \textbf{weather}                 & 97.5                                      & 97.64                                     & +0.14                                   & 97.8                                      & +0.31                                   & 97.76                                     & +0.27                                   \\ \hline
\multicolumn{2}{l}{\textbf{Overall intent accuracy}}                       & 84                                        & 84.21                                     & +0.25                                   & 89.86                                     & +6.98                                   & 90.56                                     & +7.81                                   \\ \hline
\multirow{3}{*}{\textbf{Slot F1}}         & \textbf{alarm}                   & 92.63                                     & 91.48                                     & -1.24                                   & 89.92                                     & -2.93                                   & 91.53                                     & -1.19                                   \\ \cline{2-9} 
                                          & \textbf{reminder}                & 83.49                                     & 80.64                                     & -3.41                                   & 86.09                                     & +3.11                                   & 84.5                                      & +1.21                                   \\ \cline{2-9} 
                                          & \textbf{weather}                 & 83.63                                     & 84.62                                     & +1.18                                   & 81.78                                     & -2.21                                   & 81.32                                     & -2.77                                   \\ \hline
\multicolumn{2}{l}{\textbf{Overall slot F1}}                               & 86.19                                     & 85.33                                     & -1.00                                   & 85.33                                     & -1.00                                   & 85.13                                     & -1.23                                   \\ \hline
\multicolumn{2}{l}{\textbf{Overall frame accuracy}}                        & 46.44                                     & 46.01                                     & -0.93                                   & 45.66                                     & -1.69                                   & 44.38                                     & -4.43                                   \\ \hline
\end{tabular}%
}
\end{table}

\section{Summary of the results on Language Bootstrapping Experiments averaged over three runs for Generator with token-level reward.}\label{tab:appb2}
\begin{table}[H]
\centering
%\caption{Summary of the results on Language Bootstrapping Experiments averaged over three runs for Generator with token-level reward.}
\resizebox{\textwidth}{!}{%
\begin{tabular}{llccccccccccccc}
\hline
                                           & \multicolumn{1}{c}{}                                  & \multicolumn{1}{c}{\textbf{Baseline}} & \multicolumn{6}{c}{\textbf{Generator with token-level reward}}                                                                   & \multicolumn{6}{c}{\textbf{Upsampled}}                                                                                                                                                   \\ \cline{3-15} 
                                           & \multicolumn{1}{c}{}                                  & \multicolumn{1}{c}{\textbf{}}         & \multicolumn{2}{c}{\textbf{TopX}}        & \multicolumn{2}{c}{\textbf{Uniques}}     & \multicolumn{2}{c}{\textbf{All}}         & \multicolumn{2}{c}{\textbf{TopX}}                                                                & \multicolumn{2}{c}{\textbf{Uniques}}     & \multicolumn{2}{c}{\textbf{All}}         \\ \cline{3-15} 
\multirow{-3}{*}{\textbf{Metric}}          & \multicolumn{1}{c}{\multirow{-3}{*}{\textbf{Domain}}} & \textbf{Perf.}                   & \textbf{Perf.} & \textbf{\% change} & \textbf{Perf.} & \textbf{\% change} & \textbf{Perf.} & \textbf{\% change} & \textbf{Perf.}                            & \textbf{\% change}                              & \textbf{Perf.} & \textbf{\% change} & \textbf{Perf.} & \textbf{\% change} \\ \hline
\multicolumn{2}{l}{\textbf{Domain accuracy}}                                                      & 96.83                                  & 94.08                & -2.84              & 96.49                & -0.35              & 97.11                & +0.29              & 95.41                                           & -1.47                                           & 97.05                & +0.23              & 97.50                & +0.70              \\ \hline
                                           & \textbf{alarm}                                         & 73.33                                  & 75.04                & +2.33              & 81.26                & +10.81             & 82.01                & +11.84             & 70.30                                           & -4.13                                           & 78.38                & +6.88              & 81.26                & +10.82             \\ \cline{2-15} 
                                           & \textbf{reminder}                                      & 72.93                                  & 70.64                & -3.14              & 81.40                & +11.61             & 83.60                & +14.62             & 74.30                                           & +1.87                                           & 80.07                & +9.79              & 82.47                & +13.08             \\ \cline{2-15} 
\multirow{-3}{*}{\textbf{Intent accuracy}} & \textbf{weather}                                       & 97.5                                   & 97.29                & -0.21              & 92.38                & -5.25              & 93.71                & -3.89              & 98.06                                           & +0.58                                           & 99.00                & +1.53              & 99.02                & +1.56              \\ \hline
\multicolumn{2}{l}{\textbf{Overall intent accuracy}}                                              & 84                                     & 83.33                & -0.80              & 86.26                & +2.69              & 87.70                & +4.41              & 83.40                                           & -0.71                                           & 88.06                & +4.84              & 89.64                & +6.72              \\ \hline
                                           & \textbf{alarm}                                         & 92.63                                  & 92.77                & +0.15              & 91.68                & -1.03              & 91.66                & -1.04              & 91.58                                           & -1.13                                           & 92.40                & -0.25              & 92.70                & +0.08              \\ \cline{2-15} 
                                           & \textbf{reminder}                                      & 83.49                                  & 79.36                & -4.95              & 84.23                & +0.89              & 84.16                & +0.80              & 82.28                                           & -1.45                                           & 84.48                & +1.18              & 84.87                & +1.65              \\ \cline{2-15} 
\multirow{-3}{*}{\textbf{Slot F1}}         & \textbf{weather}                                       & 83.63                                  & 84.30                & +0.80              & 84.26                & +0.75              & 84.68                & +1.25              & 83.80                                           & +0.21                                           & 83.40                & -0.27              & 82.94                & -0.83              \\ \hline
\multicolumn{2}{l}{\textbf{Overall Slot F1}}                                                      & 86.19                                  & 85.12                & -1.24              & 86.37                & +0.20              & 86.53                & +0.39              & 85.56                                           & -0.73                                           & 86.28                & +0.10              & 86.26                & +0.09              \\ \hline
\multicolumn{2}{l}{\textbf{Overall frame accuracy}}                                               & 46.44                                  & 46.61                & +0.38              & 46.81                & +0.80              & 47.95                & +3.26              & 45.47                                           & -2.08                                           & 47.86                & +3.05              & 48.47                & +4.37              \\ \hline
                                           &                                                        & \multicolumn{1}{c}{\textbf{Baseline}} & \multicolumn{6}{c}{\textbf{Generator with token-level Monte Carlo rollout}}                                                      & \multicolumn{6}{c}{\textbf{Upsampled}}                                                                                                                                                   \\ \cline{3-15} 
                                           &                                                        & \multicolumn{1}{c}{\textbf{}}         & \multicolumn{2}{c}{\textbf{TopX}}        & \multicolumn{2}{c}{\textbf{Uniques}}     & \multicolumn{2}{c}{\textbf{All}}         & \multicolumn{2}{c}{\textbf{TopX}}                                                                & \multicolumn{2}{c}{\textbf{Uniques}}     & \multicolumn{2}{c}{\textbf{All}}         \\ \cline{3-15} 
\multirow{-3}{*}{\textbf{Metric}}          & \multirow{-3}{*}{\textbf{Domain}}                      & \textbf{Perf.}                   & \textbf{Perf.} & \textbf{\% change} & \textbf{Perf.} & \textbf{\% change} & \textbf{Perf.} & \textbf{\% change} & \multicolumn{2}{c}{\textbf{{[}SAME AS ABOVE{]}}}                                                 & \textbf{Perf.} & \textbf{\% change} & \textbf{Perf.} & \textbf{\% change} \\ \hline
\multicolumn{2}{l}{\textbf{Domain accuracy}}                                                      & 96.83                                  & 94.06                & -2.86              & 97.22                & +0.41              & 94.86                & -2.03              & \cellcolor[HTML]{656565}{\color[HTML]{656565} } & \cellcolor[HTML]{656565}{\color[HTML]{656565} } & 97.81                & +1.02              & 96.18                & -0.67              \\ \hline
                                           & \textbf{alarm}                                         & 73.33                                  & 70.10                & -4.41              & 84.20                & +14.83             & 82.63                & +12.68             & \cellcolor[HTML]{656565}{\color[HTML]{656565} } & \cellcolor[HTML]{656565}{\color[HTML]{656565} } &  77.35                & +5.48              & 80.65                & +9.98              \\ \cline{2-15} 
                                           & \textbf{reminder}                                      & 72.93                                  & 71.34                & -2.18              & 83.45                & +14.43             & 77.58                & +6.38              & \cellcolor[HTML]{656565}{\color[HTML]{656565} } & \cellcolor[HTML]{656565}{\color[HTML]{656565} } & 83.39                & +14.35             & 79.49                & +9.00              \\ \cline{2-15} 
\multirow{-3}{*}{\textbf{Intent accuracy}} & \textbf{weather}                                       & 97.5                                   & 97.33                & -0.17              & 97.38                & -0.12              & 97.44                & -0.06              & \cellcolor[HTML]{656565}{\color[HTML]{656565} } & \cellcolor[HTML]{656565}{\color[HTML]{656565} } & 98.95                & +1.48              & 99.14                & +1.69              \\ \hline
\multicolumn{2}{l}{\textbf{Overall intent accuracy}}                                              & 84                                     & 81.86                & -2.55              & 90.06                & +7.21              & 87.67                & +4.36              & \cellcolor[HTML]{656565}{\color[HTML]{656565} } & \cellcolor[HTML]{656565}{\color[HTML]{656565} } & 88.64                & +5.53              & 88.61                & +5.49              \\ \hline
                                           & \textbf{alarm}                                         & 92.63                                  & 90.94                & -1.82              & 92.23                & -0.43              & 90.28                & -2.53              & \cellcolor[HTML]{656565}{\color[HTML]{656565} } & \cellcolor[HTML]{656565}{\color[HTML]{656565} } & 92.05                & -0.63              & 92.15                & -0.52              \\ \cline{2-15} 
                                           & \textbf{reminder}                                      & 83.49                                  & 79.79                & -4.43              & 85.85                & +2.82              & 81.87                & -1.94              & \cellcolor[HTML]{656565}{\color[HTML]{656565} } & \cellcolor[HTML]{656565}{\color[HTML]{656565} } & 85.45                & +2.35              & 83.94                & +0.54              \\ \cline{2-15} 
\multirow{-3}{*}{\textbf{Slot F1}}         & \textbf{weather}                                       & 83.63                                  & 85.39                & +2.10              & 83.84                & +0.25              & 84.94                & +1.56              & \cellcolor[HTML]{656565}{\color[HTML]{656565} } & \cellcolor[HTML]{656565}{\color[HTML]{656565} } & 82.62                & -1.21              & 83.56                & -0.09              \\ \hline
\multicolumn{2}{l}{\textbf{Overall Slot F1}}                                                      & 86.19                                  & 85.33                & -1.00              & 86.77                & +0.67              & 85.57                & -0.73              & \cellcolor[HTML]{656565}{\color[HTML]{656565} } & \cellcolor[HTML]{656565}{\color[HTML]{656565} } & 86.12                & -0.08              & 86.10                & -0.10              \\ \hline
\multicolumn{2}{l}{\textbf{Overall frame accuracy}}                                               & 46.44                                  & 45.36                & -2.33              & 48.44                & +4.30              & 47.75                & +2.82              & \cellcolor[HTML]{656565}{\color[HTML]{656565} } & \cellcolor[HTML]{656565}{\color[HTML]{656565} } & 47.70                & +2.71              & 48.73                & +4.92              \\ \hline
\end{tabular}}
\end{table}

%\pagebreak

\section{Summary of the results on Low-Resource Domain Experiments averaged over three runs. Pre-trained embeddings were built using robust domains and used in GAN model with token-level reward to inform the data generation for the low-resource domain.}\label{tab:appc}
\begin{table}[ht]
\centering
%\caption{Summary of the results on Low-Resource Domain Experiments averaged over three runs. Pre-trained embeddings were built using robust domains and used in GAN model with token-level reward to inform the data generation for the low-resource domain. }
\resizebox{\textwidth}{!}{%
\begin{tabular}{llccccc}
\hline
\multicolumn{7}{l}{\textbf{ALARM AS LOW-RESOURCE DOMAIN}}                                                                                                                                                                                                                \\ \hline
\multirow{2}{*}{\textbf{Metric}}          & \multirow{2}{*}{\textbf{Domain}} & \textbf{Baseline}    & \multicolumn{2}{l}{\textbf{Pre-trained embeddings with token-level reward}} & \multicolumn{2}{l}{\textbf{Pre-trained embeddings with MC rollout token-level reward}} \\ \cline{3-7} 
                                          &                                  & \textbf{Performance} & \textbf{Performance}                  & \textbf{\% change}                  & \textbf{Performance}                        & \textbf{\% change}                       \\ \hline
\multicolumn{2}{l}{\textbf{Domain accuracy}}                               & 99.19                & 99.13                                 & -0.06                               & 99.21                                       & +0.03                                    \\ \hline
\multirow{3}{*}{\textbf{Intent accuracy}} & \textbf{alarm}                   & 74.75                & 84.07                                 & +12.46                              & 85.32                                       & +14.14                                   \\ \cline{2-7} 
                                          & \textbf{reminder}                & 97.70                & 97.89                                 & +0.20                               & 97.98                                       & +0.29                                    \\ \cline{2-7} 
                                          & \textbf{weather}                 & 99.10                & 98.86                                 & -0.24                               & 99.24                                       & +0.14                                    \\ \hline
\multicolumn{2}{l}{\textbf{Overall intent accuracy}}                       & 91.56                & 94.26                                 & +2.95                               & 94.81                                       & +3.54                                    \\ \hline
\multirow{3}{*}{\textbf{Slot F1}}         & \textbf{alarm}                   & 93.00                & 91.34                                 & -1.79                               & 92.46                                       & -0.58                                    \\ \cline{2-7} 
                                          & \textbf{reminder}                & 92.79                & 92.70                                 & -0.10                               & 92.79                                       & +0.00                                    \\ \cline{2-7} 
                                          & \textbf{weather}                 & 97.79                & 97.78                                 & -0.01                               & 97.86                                       & +0.07                                    \\ \hline
\multicolumn{2}{l}{\textbf{Overall Slot F1}}                               & 95.13                & 94.64                                 & -0.52                               & 95.00                                       & -0.13                                    \\ \hline
\multicolumn{2}{l}{\textbf{Overall frame accuracy}}                        & 73.95                & 74.14                                 & +0.27                               & 74.57                                       & +0.84                                    \\ \hline
\multicolumn{7}{l}{\textbf{REMINDER AS LOW-RESOURCE DOMAIN}}                                                                                                                                                                                                             \\ \hline
\multirow{2}{*}{\textbf{Metric}}          & \multirow{2}{*}{\textbf{Domain}} & \textbf{Baseline}    & \multicolumn{2}{l}{\textbf{Pre-trained embeddings with token-level reward}} & \multicolumn{2}{l}{\textbf{Pre-trained embeddings with MC rollout token-level reward}} \\ \cline{3-7} 
                                          &                                  & \textbf{Performance} & \textbf{Performance}                  & \textbf{\% change}                  & \textbf{Performance}                        & \textbf{\% change}                       \\ \hline
\multicolumn{2}{l}{\textbf{Domain accuracy}}                               & 99.68                & 99.57                                 & -0.12                               & 99.40                                       & -0.28                                    \\ \hline
\multirow{3}{*}{\textbf{Intent accuracy}} & \textbf{alarm}                   & 96.75                & 96.80                                 & +0.05                               & 96.59                                       & -0.16                                    \\ \cline{2-7} 
                                          & \textbf{reminder}                & 80.81                & 91.48                                 & +13.21                              & 93.24                                       & +15.39                                   \\ \cline{2-7} 
                                          & \textbf{weather}                 & 99.75                & 99.57                                 & -0.18                               & 99.37                                       & -0.38                                    \\ \hline
\multicolumn{2}{l}{\textbf{Overall intent accuracy}}                       & 94.56                & 96.90                                 & +2.47                               & 97.15                                       & +2.74                                    \\ \hline
\multirow{3}{*}{\textbf{Slot F1}}         & \textbf{alarm}                   & 96.29                & 96.32                                 & +0.03                               & 96.24                                       & -0.05                                    \\ \cline{2-7} 
                                          & \textbf{reminder}                & 87.90                & 87.63                                 & -0.31                               & 88.00                                       & +0.11                                    \\ \cline{2-7} 
                                          & \textbf{weather}                 & 97.96                & 97.83                                 & -0.14                               & 97.84                                       & -0.13                                    \\ \hline
\multicolumn{2}{l}{\textbf{Overall Slot F1}}                               & 94.94                & 94.82                                 & -0.13                               & 94.92                                       & -0.02                                    \\ \hline
\multicolumn{2}{l}{\textbf{Overall frame accuracy}}                        & 75.70                & 76.05                                 & +0.46                               & 76.16                                       & +0.61                                    \\ \hline
\multicolumn{7}{l}{\textbf{WEATHER AS LOW-RESOURCE DOMAIN}}                                                                                                                                                                                                              \\ \hline
\multirow{2}{*}{\textbf{Metric}}          & \multirow{2}{*}{\textbf{Domain}} & \textbf{Baseline}    & \multicolumn{2}{l}{\textbf{Pre-trained embeddings with token-level reward}} & \multicolumn{2}{l}{\textbf{Pre-trained embeddings with MC rollout token-level reward}} \\ \cline{3-7} 
                                          &                                  & \textbf{Performance} & \textbf{Performance}                  & \textbf{\% change}                  & \textbf{Performance}                        & \textbf{\% change}                       \\ \hline
\multicolumn{2}{l}{\textbf{Domain accuracy}}                               & 96.64                & 95.46                                 & -1.22                               & 96.55                                       & -0.09                                    \\ \hline
\multirow{3}{*}{\textbf{Intent accuracy}} & \textbf{alarm}                   & 92.66                & 89.94                                 & -2.93                               & 91.64                                       & -1.10                                    \\ \cline{2-7} 
                                          & \textbf{reminder}                & 92.30                & 91.63                                 & -0.72                               & 93.30                                       & +1.08                                    \\ \cline{2-7} 
                                          & \textbf{weather}                 & 97.73                & 97.80                                 & +0.08                               & 98.23                                       & +0.51                                    \\ \hline
\multicolumn{2}{l}{\textbf{Overall intent accuracy}}                       & 94.75                & 93.64                                 & -1.18                               & 94.86                                       & +0.12                                    \\ \hline
\multirow{3}{*}{\textbf{Slot F1}}         & \textbf{alarm}                   & 94.57                & 92.92                                 & -1.74                               & 93.75                                       & -0.86                                    \\ \cline{2-7} 
                                          & \textbf{reminder}                & 89.25                & 88.82                                 & -0.49                               & 90.09                                       & +0.94                                    \\ \cline{2-7} 
                                          & \textbf{weather}                 & 84.72                & 84.36                                 & -0.42                               & 83.13                                       & -1.87                                    \\ \hline
\multicolumn{2}{l}{\textbf{Overall Slot F1}}                               & 89.02                & 88.32                                 & -0.78                               & 88.38                                       & -0.71                                    \\ \hline
\multicolumn{2}{l}{\textbf{Overall frame accuracy}}                        & 60.38                & 59.04                                 & -2.22                               & 58.29                                       & -3.46                                    \\ \hline
\end{tabular}%
}
\end{table}

\section{Data quality evaluation results.}\label{tab:appd}
\begin{table}[h]
\centering
%\caption{Data quality evaluation results.}
\resizebox{\textwidth}{!}{%
\begin{tabular}{lllccccccc}
\hline
\textbf{Model}                                                                                                                     & \textbf{\begin{tabular}[c]{@{}l@{}}Pre-trained \\ embeddings\end{tabular}} & \textbf{Domain}   & \textbf{Unique utterances} & \textbf{Unique words} & \textbf{Mean utterance length} & \textbf{BLEU1} & \textbf{BLEU2} & \textbf{BLEU3} & \textbf{BLEU4} \\ \hline
\multirow{3}{*}{\textbf{\begin{tabular}[c]{@{}l@{}}Generator with sentence-level \\ reward without rollouts\end{tabular}}}         & \multirow{3}{*}{\textbf{No}}                                               & \textbf{alarm}    & 1183                       & 113                   & 5.49                           & 96.76          & 91.23          & 82.38          & 61.45          \\ \cline{3-10} 
                                                                                                                                   &                                                                            & \textbf{reminder} & 2792                       & 539                   & 6.23                           & 94.64          & 85.51          & 73.44          & 60.9           \\ \cline{3-10} 
                                                                                                                                   &                                                                            & \textbf{weather}  & 1378                       & 104                   & 5.33                           & 99.77          & 96.03          & 90.78          & 81.44          \\ \hline
\multirow{3}{*}{\textbf{\begin{tabular}[c]{@{}l@{}}Generator with sentence-level \\ reward with Monte Carlo rollout\end{tabular}}} & \multirow{3}{*}{\textbf{No}}                                               & \textbf{alarm}    & 1064                       & 111                   & 5.73                           & 97.91          & 94.37          & 88.72          & 77.1           \\ \cline{3-10} 
                                                                                                                                   &                                                                            & \textbf{reminder} & 3487                       & 540                   & 6.75                           & 95.1           & 84.83          & 70.99          & 53.93          \\ \cline{3-10} 
                                                                                                                                   &                                                                            & \textbf{weather}  & 1839                       & 105                   & 5.59                           & 99.65          & 91.98          & 85.06          & 75.22          \\ \hline
\multirow{3}{*}{\textbf{\begin{tabular}[c]{@{}l@{}}Generator with \\ token-level reward\end{tabular}}}                             & \multirow{3}{*}{\textbf{No}}                                               & \textbf{alarm}    & 2350                       & 122                   & 6.24                           & 94.7           & 87.49          & 78.17          & 66.12          \\ \cline{3-10} 
                                                                                                                                   &                                                                            & \textbf{reminder} & 4729                       & 580                   & 6.78                           & 98.97          & 85.71          & 69.35          & 54.98          \\ \cline{3-10} 
                                                                                                                                   &                                                                            & \textbf{weather}  & 1479                       & 108                   & 5.45                           & 99.62          & 96.13          & 91.16          & 86.33          \\ \hline
\multirow{3}{*}{\textbf{\begin{tabular}[c]{@{}l@{}}Generator with token-level \\ Monte Carlo rollout\end{tabular}}}                & \multirow{3}{*}{\textbf{No}}                                               & \textbf{alarm}    & 2123                       & 123                   & 6.11                           & 96.29          & 90.32          & 75.79          & 62.03          \\ \cline{3-10} 
                                                                                                                                   &                                                                            & \textbf{reminder} & 5144                       & 593                   & 6.52                           & 99.01          & 87.03          & 67.57          & 49.29          \\ \cline{3-10} 
                                                                                                                                   &                                                                            & \textbf{weather}  & 1975                       & 109                   & 5.44                           & 99.22          & 95.09          & 87.75          & 76.11          \\ \hline
\multirow{3}{*}{\textbf{\begin{tabular}[c]{@{}l@{}}Generator with \\ token-level reward\end{tabular}}}                             & \multirow{3}{*}{\textbf{Yes}}                                              & \textbf{alarm}    & 2545                       & 116                   & 6.07                           & 94.59          & 83.96          & 75.01          & 65.82          \\ \cline{3-10} 
                                                                                                                                   &                                                                            & \textbf{reminder} & 7048                       & 572                   & 7.07                           & 98.03          & 79.2           & 58.34          & 41.35          \\ \cline{3-10} 
                                                                                                                                   &                                                                            & \textbf{weather}  & 1323                       & 104                   & 5.59                           & 99.64          & 89.51          & 83.89          & 74.8           \\ \hline
\multirow{3}{*}{\textbf{\begin{tabular}[c]{@{}l@{}}Generator with token-level \\ Monte Carlo rollout\end{tabular}}}                & \multirow{3}{*}{\textbf{Yes}}                                              & \textbf{alarm}    & 3451                       & 112                   & 7.22                           & 91.45          & 80.24          & 69.59          & 56.38          \\ \cline{3-10} 
                                                                                                                                   &                                                                            & \textbf{reminder} & 6128                       & 586                   & 6.58                           & 96.08          & 78.14          & 58.32          & 39.87          \\ \cline{3-10} 
                                                                                                                                   &                                                                            & \textbf{weather}  & 2291                       & 108                   & 5.43                           & 98.08          & 87.39          & 75.49          & 58.32          \\ \hline
\end{tabular}%
}
\end{table}

\end{document}